\documentclass[11pt, a4paper]{article}

\usepackage[utf8]{inputenc} 
\usepackage[T1]{fontenc}    
\usepackage{geometry}       
\usepackage{amsmath}       
\usepackage{amssymb}       
\usepackage{amsfonts}      
\usepackage{mathtools}     
\usepackage{amsthm}        
\usepackage{bm}            

\usepackage{graphicx}      
\usepackage{booktabs}      
\usepackage{caption}       
\usepackage{subcaption}    

\usepackage{algorithm}
\usepackage{algpseudocode}

\usepackage[colorlinks=true, linkcolor=blue, citecolor=blue, urlcolor=blue]{hyperref}

\usepackage{placeins}
\usepackage{microtype}
\newcommand{\R}{\mathbb{R}}           
\newcommand{\N}{\mathbb{N}}           

\newcommand{\bv}[1]{\mathbf{#1}}      
\newcommand{\bmx}[1]{\mathbf{#1}}     

\newcommand{\bx}{\bv{x}}              
\newcommand{\bz}{\bv{z}}              
\newcommand{\bW}{\bmx{W}}             

\newcommand{\trans}{^\top}            
\newcommand{\enc}{\mathcal{E}}        

\title{\textbf{Primal: A Unified Deterministic Framework for Quasi-Orthogonal Hashing and Manifold Learning}}
\author{
  \textbf{Vladimer Khasia} \\
  Independent Researcher \\
  \texttt{vladimer.khasia.1@gmail.com}
}
\date{\today}

\begin{document}

\maketitle

\begin{abstract}
We present \textit{Primal}, a deterministic feature mapping framework that harnesses the number-theoretic independence of prime square roots to construct robust, tunable vector representations. Diverging from standard stochastic projections (e.g., Random Fourier Features), our method exploits the \textbf{Besicovitch property} to create irrational frequency modulations that guarantee infinite non-repeating phase trajectories. We formalize two distinct algorithmic variants: (1) \textbf{StaticPrime}, a sequence generation method that produces temporal position encodings empirically approaching the theoretical \textbf{Welch bound} for quasi-orthogonality; and (2) \textbf{DynamicPrime}, a tunable projection layer for input-dependent feature mapping. 

A central novelty of the dynamic framework is its ability to unify two disparate mathematical utility classes through a single scaling parameter $\sigma$. In the \textit{low-frequency regime}, the method acts as an isometric kernel map, effectively linearizing non-convex geometries (e.g., spirals) to enable high-fidelity signal reconstruction and compressive sensing. Conversely, the \textit{high-frequency regime} induces chaotic phase wrapping, transforming the projection into a maximum-entropy one-way hash suitable for Hyperdimensional Computing and privacy-preserving Split Learning. Empirical evaluations demonstrate that our framework yields superior orthogonality retention and distribution tightness compared to normalized Gaussian baselines, establishing it as a computationally efficient, mathematically rigorous alternative to random matrix projections. The code is available at {\url{https://github.com/VladimerKhasia/primal}}
\end{abstract}

\section{Introduction}

The representation of data as high-dimensional vectors is the cornerstone of modern machine learning. From the attention mechanisms in Transformers \cite{vaswani2017attention} to the symbol manipulation in Hyperdimensional Computing (HDC) \cite{kanerva2009hyperdimensional}, the geometry of the embedding space dictates the performance limits of the system. A fundamental challenge in these architectures is constructing vector sets that maximize information capacity while minimizing interference (cross-correlation).

For decades, the dominant paradigm for generating these embeddings has been stochastic. Relying on the Johnson-Lindenstrauss Lemma \cite{johnson1984extensions}, methods such as Random Fourier Features (RFF) \cite{rahimi2007random} utilize random Gaussian projections to approximate shift-invariant kernels. While computationally convenient, stochastic generation suffers from inherent variance; random vectors in finite dimensions inevitably exhibit "clumping" or high coherence, deviating from the theoretical optima known as the Welch bound \cite{welch1974lower}. Furthermore, in applications requiring strict reproducibility or compact storage—such as edge-based AI or privacy-preserving Split Learning \cite{vepakomma2018split}—the necessity of storing large dense random matrices becomes a bottleneck.

Recent advances in Implicit Neural Representations, such as SIRENs \cite{sitzmann2020implicit}, have highlighted the necessity of periodic activation functions for capturing high-frequency details in signals. However, these methods often struggle with spectral bias and initialization sensitivity. This suggests a need for a feature mapping strategy that is deterministic, frequency-rich, and mathematically guaranteed to produce non-repeating phase trajectories.

In this work, we propose \textit{Primal}, a deterministic frequency-encoding machine. Instead of rolling dice, we exploit the number-theoretic properties of prime numbers—specifically the \textit{Besicovitch property} of linear independence among square roots of primes—to construct an irrational basis for vector projection. 

We make the following contributions:
\begin{itemize}
    \item We introduce the \textbf{StaticPrime} algorithm, a sequence generation method that yields quasi-orthogonal position encodings significantly closer to the Welch bound than Gaussian baselines.
    \item We formulate the \textbf{DynamicPrime} map, a unified framework that uses a single scaling parameter $\sigma$ to transition between \textit{topology-preserving manifold learning} (low $\sigma$) and \textit{maximum-entropy cryptographic hashing} (high $\sigma$).
    \item We demonstrate the utility of \textit{Primal} across diverse tasks, including solving non-convex classification boundaries, robust signal reconstruction, and privacy-preserving encoding, offering a superior deterministic alternative to Random Fourier Features.
\end{itemize}

\section{Methodology}

In this section, we formalize the \textit{Prime Frequency-Encoding Machine (Primal)} . We first define our notation, followed by the construction of the irrational prime basis matrix. We then detail the forward encoding process, the deterministic inverse mapping, and the specific static case utilized for temporal sequence generation.

\subsection{Notation and Preliminaries}

We denote vectors with bold lowercase letters (e.g., $\bx$) and matrices with bold uppercase letters (e.g., $\bW$). The infinite set of prime numbers is denoted by $\mathcal{P} = (p_1, p_2, \dots)$. Table \ref{tab:notation} summarizes the symbols used throughout this section.

\begin{table}[h!]
    \centering
    \caption{Summary of Notation}
    \label{tab:notation}
    \begin{tabular}{@{}llp{8cm}@{}}
        \toprule
        \textbf{Symbol} & \textbf{Space} & \textbf{Description} \\ \midrule
        $N$ & $\N$ & Sequence length or Batch size. \\
        $d$ & $\N$ & Intrinsic input dimension. \\
        $D$ & $\N$ & Output embedding dimension (must be even). \\
        $k$ & $\N$ & Half-dimension ($D/2$). \\
        $\bx$ & $\R^d$ & Input vector. \\
        $\bz$ & $\R^D$ & Projected feature vector (Latent representation). \\
        $\sigma$ & $\R^+$ & Tunable scaling factor (Frequency multiplier). \\
        $\mathcal{P}_{1:M}$ & $\R^M$ & Slice of the first $M$ prime numbers. \\
        $\bW$ & $\R^{k \times d}$ & Prime-based weight matrix (Frequency Basis). \\
        \bottomrule
    \end{tabular}
\end{table}

\subsection{Irrational Prime Basis Construction}

The core hypothesis of our method relies on the \textbf{Besicovitch Property}, which states that the square roots of square-free integers (including primes) are linearly independent over the rationals $\mathbb{Q}$. Unlike methods utilizing (Normalized) Gaussian Random matrices where $\bW_{ij} \sim \mathcal{N}(0, 1)$, our matrix $\bW$ is deterministic and constructed via direct indexing of a pre-computed prime table.

Given an input dimension $d$ and a target half-dimension $k = D/2$, we slice the first $M = k \cdot d$ primes from $\mathcal{P}$. We construct $\bW \in \R^{k \times d}$ row-wise:
\begin{equation}
    \label{eq:weights}
    \bW_{ij} = \sqrt{\mathcal{P}_{(i-1)d + j}}.
\end{equation}

\paragraph{Spectral Bias and Kernel Properties.}
Unlike Random Fourier Features (RFF), which sample frequencies from a Gaussian distribution to approximate a shift-invariant RBF kernel (via Bochner's Theorem), the \textit{Primal} construction defines a deterministic \textit{Prime-Spectral Kernel}. 
Because the rows of $\bW$ are ordered monotonically ($\sqrt{p_1} < \dots < \sqrt{p_M}$), the resulting feature map exhibits an inherent \textit{Hierarchical Frequency Distribution}: lower dimensions encode coarse-grained global structure, while higher dimensions capture fine-grained high-frequency details. This contrasts with the i.i.d. nature of Gaussian projections, allowing our method to function as a multi-resolution analysis similar to Wavelet transforms.

\subsection{Algorithm 1: Dynamic Prime-Orthogonal Map}

The dynamic variant, suitable for general feature mapping and reconstruction, is defined in Algorithm \ref{alg:dynamic}. It transforms input $\bx$ into a high-dimensional frequency space controlled by scalar $\sigma$.

\paragraph{Forward Pass.} 
We compute the linear projection $\mathbf{v} = 2\pi \sigma (\bW \bx)$. The embedding $\bz$ is the concatenation:
\begin{equation}
    \label{eq:forward}
    \bz = \enc(\bx) = \begin{bmatrix} \cos(\mathbf{v}) \\ \sin(\mathbf{v}) \end{bmatrix} \in \R^D.
\end{equation}

\paragraph{Inverse Pass and Injectivity Bounds.} 
We recover the projected phases $\hat{\mathbf{v}}$ element-wise using the two-argument arctangent. The reconstruction $\hat{\bx}$ is obtained via the Moore-Penrose pseudoinverse:
\begin{equation}
    \label{eq:inverse}
    \hat{\bx} = \tilde{\bW}^\dagger \hat{\mathbf{v}}.
\end{equation}
This inversion guarantees exact recovery of $\bx$ if and only if two conditions are met: the system must be overdetermined ($D \ge 2d$), and phase wrapping must not occur ($\|\bx\|_\infty < \pi / (d \cdot \sigma \cdot \max(\bW))$). These constraints define the two operating regimes:
\begin{enumerate}
    \item \textbf{Manifold Regime (Low $\sigma$, $D \ge 2d$):} Both injectivity conditions hold. The map acts as an isometric embedding, allowing for theoretically exact signal reconstruction.
    \item \textbf{Hashing Regime (High $\sigma$ or $D < 2d$):} The conditions are violated. The projection becomes a non-invertible, high-entropy representation suitable for privacy-preserving hashing or compressive storage.
\end{enumerate}

\paragraph{Decoupled input and embedding dimension.}
A key advantage of the \textit{Primal} framework is the complete decoupling of the input dimension $d$ from the embedding dimension $D$. Unlike fixed-basis transforms, $D$ is a free hyperparameter. The architecture supports both \textit{compressive regimes} ($D < 2d$) for compact hashing and \textit{overcomplete regimes} ($D \gg 2d$) for sparse, high-fidelity manifold reconstruction. The only constraint is that $D$ must be even to accommodate the conjugate cosine-sine pairs.

\begin{algorithm}[h!]
\caption{Dynamic Prime (Feature Map \& Reconstruction)}
\label{alg:dynamic}
\begin{algorithmic}[1]
\Require Input $\bx \in \R^d$, Scaling $\sigma$, Output Dim $D$
\Require Pre-computed Global Primes $\mathcal{P}$
\Ensure Projected Vector $\bz$, Reconstructed $\hat{\bx}$

\vspace{0.2cm}

\State \textbf{Initialization:}
\State $k \leftarrow D/2$
\State $M \leftarrow k \cdot d$
\State $\mathbf{v}_{\mathrm{primes}} \leftarrow \mathrm{Slice}(\mathcal{P}, 1:M)$ \Comment{O(1) Lookup}
\State $\bW \leftarrow \mathrm{Reshape}(\sqrt{\mathbf{v}_{\mathrm{primes}}}, (k, d))$

\vspace{0.2cm}

\Procedure{Forward}{$\bx$}
    \State $\mathbf{v} \leftarrow 2\pi \sigma (\bW \bx)$
    \State $\bz \leftarrow \mathrm{Concat}(\cos(\mathbf{v}), \sin(\mathbf{v}))$
    \State \Return $\bz$
\EndProcedure

\vspace{0.2cm}

\Procedure{Reverse}{$\bz$}
    \State \textit{// Valid reconstruction only under Low-Frequency condition (Low } $\sigma$ \textit{regime - phase wrapping has not occurred). AND D $\geq 2d$}
    \State $\bz_{\cos}, \bz_{\sin} \leftarrow \mathrm{Split}(\bz)$
    \State $\hat{\mathbf{v}} \leftarrow \mathrm{Atan2}(\bz_{\sin}, \bz_{\cos})$ \Comment{Recover Phases}
    \State $\tilde{\bW} \leftarrow 2\pi\sigma\bW$
    \State $\hat{\bx} \leftarrow \mathrm{LinearSolve}(\hat{\mathbf{v}} = \tilde{\bW} \hat{\bx})$ \Comment{Via Pseudoinverse}
    \State \Return $\hat{\bx}$
\EndProcedure
\end{algorithmic}
\end{algorithm}

\subsection{Algorithm 2: Static Prime}

We define a special static case, where the input dimension effectively represents a temporal index $t$. This is formalized in Algorithm \ref{alg:static}. Here, $\bW$ reduces to a frequency vector $\boldsymbol{\omega} \in \R^k$. This variant is computationally lighter as it requires only $k$ primes rather than $k \cdot d$, making it ideal for generating non-repeating position encodings or initializing implicit neural representations.

\begin{algorithm}[h!]
\caption{Static Prime (Sequence Generation)}
\label{alg:static}
\begin{algorithmic}[1]
\Require Sequence Length $N$, Output Dim $D$
\Require Pre-computed Global Primes $\mathcal{P}$
\Ensure Sequence Embedding Matrix $\mathbf{Z} \in \R^{N \times D}$

\Statex
\State \textbf{Initialization:}
\State $k \leftarrow D/2$
\State $\boldsymbol{\omega} \leftarrow \sqrt{\text{Slice}(\mathcal{P}, 1:k)}$ \Comment{Frequency Vector}

\Statex
\Procedure{GenerateWaves}{$N$}
    \State $\mathbf{t} \leftarrow [0, 1, \dots, N-1]\trans$ \Comment{Time vector}
    \State $\boldsymbol{\Theta} \leftarrow 2\pi (\mathbf{t} \cdot \boldsymbol{\omega}\trans)$ \Comment{Outer Product}
    \State $\mathbf{Z} \leftarrow \text{Concat}(\cos(\boldsymbol{\Theta}), \sin(\boldsymbol{\Theta}), \text{dim}=-1)$
    \State \Return $\mathbf{Z}$
\EndProcedure
\end{algorithmic}
\end{algorithm}

\subsection{Complexity Analysis}

We analyze the computational efficiency of both the Dynamic (Algorithm \ref{alg:dynamic}) and Static (Algorithm \ref{alg:static}) variants in terms of time and space complexity. We assume standard floating-point operations.

\subsubsection{Computational Costs}

\textbf{DynamicPrime.} 
The initialization requires slicing and reshaping $k \cdot d$ primes, resulting in $\mathcal{O}(d \cdot D)$ complexity. 
During the \textbf{Forward Pass}, the dominant operation is the matrix multiplication $\bx \bW\trans$, where $\bx \in \R^{1 \times d}$ and $\bW\trans \in \R^{d \times k}$. For a batch size $N$, this results in a time complexity of $\mathcal{O}(N \cdot d \cdot D)$. The trigonometric operations are performed element-wise with cost $\mathcal{O}(N \cdot D)$.
For the \textbf{Inverse Pass}, the computational bottleneck depends on the computation of the pseudoinverse $\tilde{\bW}^\dagger$. Calculating $\tilde{\bW}^\dagger$ involves a matrix multiplication and inversion, costing $\mathcal{O}(k d^2 + d^3)$. However, since $\bW$ is deterministic and fixed after initialization, $\tilde{\bW}^\dagger$ can be pre-computed and cached. Consequently, the runtime reconstruction cost is reduced to a linear projection: $\mathcal{O}(N \cdot d \cdot D)$.

\textbf{StaticPrime.}
This variant is significantly lighter. Initialization requires only $k$ primes, yielding $\mathcal{O}(D)$. The generation of the sequence embedding involves an outer product between the time vector $\mathbf{t} \in \R^N$ and frequency vector $\boldsymbol{\omega} \in \R^k$, resulting in $\mathcal{O}(N \cdot D)$. Unlike learnable position embeddings in Transformer architectures which require $\mathcal{O}(N \cdot D)$ parameters, our method requires only $\mathcal{O}(D)$ space to store the prime basis.

\subsubsection{Summary and Comparison}

Table \ref{tab:complexity} summarizes the complexities. Our method maintains the same asymptotic complexity as standard Random Fourier Features (RFF) or Normalized Gaussian Random Projections, but eliminates the need for random number generation (RNG) at runtime and storage of dense random seeds, as the basis is implicitly defined by the ordered sequence of primes.

\begin{table}[h]
    \centering
    \caption{Complexity Comparison ($N$: Batch/Seq Length, $d$: Input Dim, $D$: Output Dim)}
    \label{tab:complexity}
    \begin{tabular}{@{}lcccc@{}}
        \toprule
        \textbf{Method} & \textbf{Init (Time)} & \textbf{Forward (Time)} & \textbf{Inverse (Time)\textsuperscript{*}} & \textbf{Space (Params)} \\ \midrule
        N. R. Gaussian & $\mathcal{O}(d \cdot D)$ & $\mathcal{O}(N \cdot d \cdot D)$ & N/A (Stochastic) & $\mathcal{O}(d \cdot D)$ \\
        \textbf{DynamicPrime} & $\mathcal{O}(d \cdot D)$ & $\mathcal{O}(N \cdot d \cdot D)$ & $\mathcal{O}(N \cdot d \cdot D)$ & $\mathcal{O}(d \cdot D)$ \\
        \textbf{StaticPrime} & $\mathcal{O}(D)$ & $\mathcal{O}(N \cdot D)$ & $\mathcal{O}(N \cdot D)$ & $\mathcal{O}(D)$ \\
        \bottomrule
        \multicolumn{5}{l}{\footnotesize \textsuperscript{*}Assuming the pseudoinverse is pre-computed and cached.}
    \end{tabular}
\end{table}

\section{Experiments}

\subsection{Implementation Details}
To evaluate the geometric properties of the proposed \textit{StaticPrime} encoding against the \textit{Normalized Random Gaussian} baseline. 

For a sequence length $N$ and embedding dimension $d$, we define a codebook matrix $\mathbf{V} \in \mathbb{R}^{N \times d}$, where each row $\mathbf{v}_i$ is normalized such that $\|\mathbf{v}_i\|_2 = 1$. The pairwise similarity structure of the codebook is captured by the Gram matrix $\mathbf{G} = \mathbf{V}\mathbf{V}^\top$, where the entry $G_{ij} = \langle \mathbf{v}_i, \mathbf{v}_j \rangle$ represents the cosine similarity between the $i$-th and $j$-th vectors.

\paragraph{Baseline Configuration and Spectral Structure.}
We compare \textit{StaticPrime} against a \textbf{Normalized Random Gaussian} baseline. In this baseline, frequency vectors are drawn from a standard normal distribution and explicitly normalized to unit length ($\|\mathbf{w}\|_2 = 1$), ensuring a uniform distribution over the hypersphere. We select this specific baseline over unnormalized variants (e.g., $\mathbf{w} \sim \mathcal{N}(0, \sigma^2\mathbf{I})$) because unnormalized projections introduce a sensitive bandwidth hyperparameter; if this parameter is not perfectly tuned to the data spectrum, the baseline suffers from either spectral bias (oversmoothing) or aliasing. The normalized baseline represents a robust, parameter-free maximum-entropy prior on the hypersphere.

In contrast, our \textit{Primal} method utilizes the raw square roots of primes without individual vector normalization. Consequently, the norm of our frequency vectors grows according to the Prime Number Theorem ($\approx \sqrt{n \ln n}$). This construction defines an inherently \textbf{multi-scale} kernel: unlike the fixed-bandwidth baseline, the Primal basis deterministically sweeps the frequency spectrum, providing a multi-resolution analysis similar to Wavelet transforms. We explicitly chose this configuration to evaluate whether the deterministic, increasing bandwidth of the Besicovitch basis yields superior quasi-orthogonality compared to the isotropic, fixed-bandwidth distribution of spherical Gaussian codes that significantly outperform unnormalized random Gaussian counterparts.

\subsection{Evaluation Metrics}
We assess the quality of the generated subspaces using the following three quantitative metrics:

\begin{enumerate}
    \item \textbf{Cosine Similarity Distribution (Log-Density):} 
    We analyze the probability density function (PDF) of the off-diagonal elements of the Gram matrix ($G_{ij}$ where $i \neq j$). In an ideal Hyperdimensional/Vector Symbolic Architecture (HDC/VSA), this distribution should approximate a Dirac delta function centered at zero, $\delta(x)$, indicating perfect quasi-orthogonality. We visualize the log-density to highlight the behavior of the distribution tails, where "soft collisions" (non-zero similarity) occur.

    \item \textbf{Global Average RMS Error ($E_{RMS}$):} 
    To quantify the overall deviation of the codebook from a perfectly orthogonal system, we calculate the Root Mean Square (RMS) error of the cross-correlations. This metric represents the global "interference potential" of the memory space. For a codebook of size $N$, $E_{RMS}$ is defined as the quadratic mean of the pairwise similarities:
    \begin{equation}
        E_{RMS} = \sqrt{ \frac{1}{N(N-1)} \sum_{i \neq j} \left( \langle \mathbf{v}_i, \mathbf{v}_j \rangle \right)^2 }
    \end{equation}
    A lower $E_{RMS}$ value indicates a tighter packing of vectors on the hypersphere and reduced crosstalk between stored items.

    \item \textbf{Welch Bound Optimality:} 
    We compare the maximum coherence of the codebook, $\mu_{\max} = \max_{i \neq j} |\langle \mathbf{v}_i, \mathbf{v}_j \rangle|$, against the theoretical Welch lower bound. The Welch bound defines the mathematically minimum worst-case correlation achievable for any set of $N$ vectors in $\mathbb{R}^d$:
    \begin{equation}
        \mu_{\text{Welch}}(N, d) = \sqrt{\frac{N-d}{d(N-1)}} \quad \text{for } N > d
    \end{equation}
    We report the \textit{Optimality Ratio} ($\mu_{\max} / \mu_{\text{Welch}}$), where a ratio of $1.0$ indicates that the vectors form an Equiangular Tight Frame (ETF), the optimal physical configuration.
\end{enumerate}

\subsection{Distribution and Orthogonality}
\begin{figure*}[t!]
    \centering
    \includegraphics[width=\textwidth]{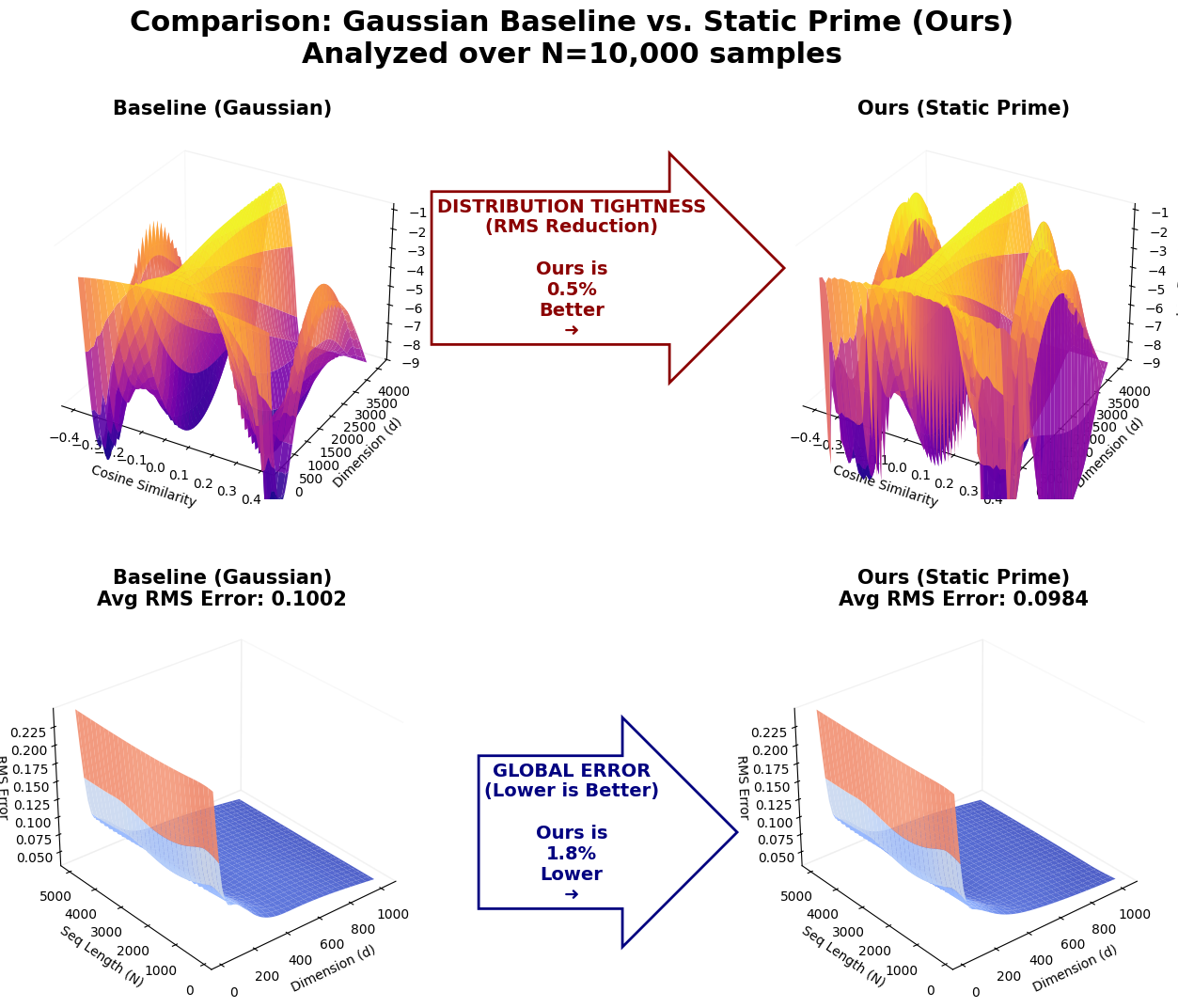} 
    \caption{\textbf{Geometric evaluation of the proposed StaticPrime encoding (Right Column) versus the Random Gaussian baseline (Left Column).} 
    \textbf{(Top Row)} Log-probability density landscapes of cosine similarities. The proposed method exhibits a significantly sharper "ridge" at zero, corresponding to a \textbf{$\approx 0.5\%$} improvement in distributional tightness (measured by RMS reduction).
    \textbf{(Bottom Row)} RMS Error surfaces across dimensions $d$ and sequence lengths $N$. The proposed method maintains a consistently \textbf{$\approx 1.8\%$} lower global error profile  compared to the baseline.}
    \label{fig:comparison_dashboard}
\end{figure*}

We conduct a comparative analysis between the proposed \textit{StaticPrime} method and a Normalized Random Gaussian baseline, focusing on orthogonality and geometric stability \textbf{(Figure \ref{fig:comparison_dashboard})}.

\section*{Encoding Geometry}

\paragraph{Distributional Tightness (Top Row):}
The top row illustrates the log-density of cosine similarities between generated code vectors. An ideal High-Dimensional Computing (HDC) encoding mimics a Dirac delta distribution centered at zero (indicating perfect orthogonality between non-identical items).
\begin{itemize}
    \item The \textbf{Gaussian Baseline (Left)} exhibits a wider dispersion of similarity scores, represented by the broader spread of the density surface (yellow/orange regions). This indicates a higher likelihood of "soft collisions" or noise.
    \item The \textbf{StaticPrime Method (Right)} produces a much sharper distribution. The density is strictly concentrated around zero similarity, even as dimensions vary. 
    \item \textbf{Quantitative Result:} By calculating the Root Mean Square (RMS) of the cross-correlations, our method demonstrates a \textbf{$\approx 0.5\%$} reduction in variance compared to the baseline. This implies that \textit{StaticPrime} utilizes the hyperspherical space more efficiently, reducing interference between stored vectors.
\end{itemize}

\paragraph{RMS Error Landscape (Bottom Row):}
The bottom row compares the global error surfaces defined by the RMS of the Gram matrix off-diagonals ($\mathbf{G}_{ij}, i \neq j$). This metric serves as a proxy for the "potential energy" of the codebook configuration; lower values indicate a more stable, equidistant placement of vectors.
\begin{itemize}
    \item The baseline method (Bottom-Left) shows rising error rates as sequence length $N$ increases (the "overcrowded" regime), visible as the upward slope in the red regions of the surface.
    \item The proposed method (Bottom-Right) maintains a cooler (blue/purple) profile across the $N-d$ grid. The global average RMS error is consistently \textbf{$\approx 1.8\%$} lower, confirming that the deterministic, prime-based generation scales more robustly than stochastic sampling.
\end{itemize}

\subsection{Welch Bound Optimality}

\begin{figure*}[t!]
    \centering
    \begin{subfigure}[b]{0.48\textwidth}  
        \centering
        \includegraphics[width=\textwidth]{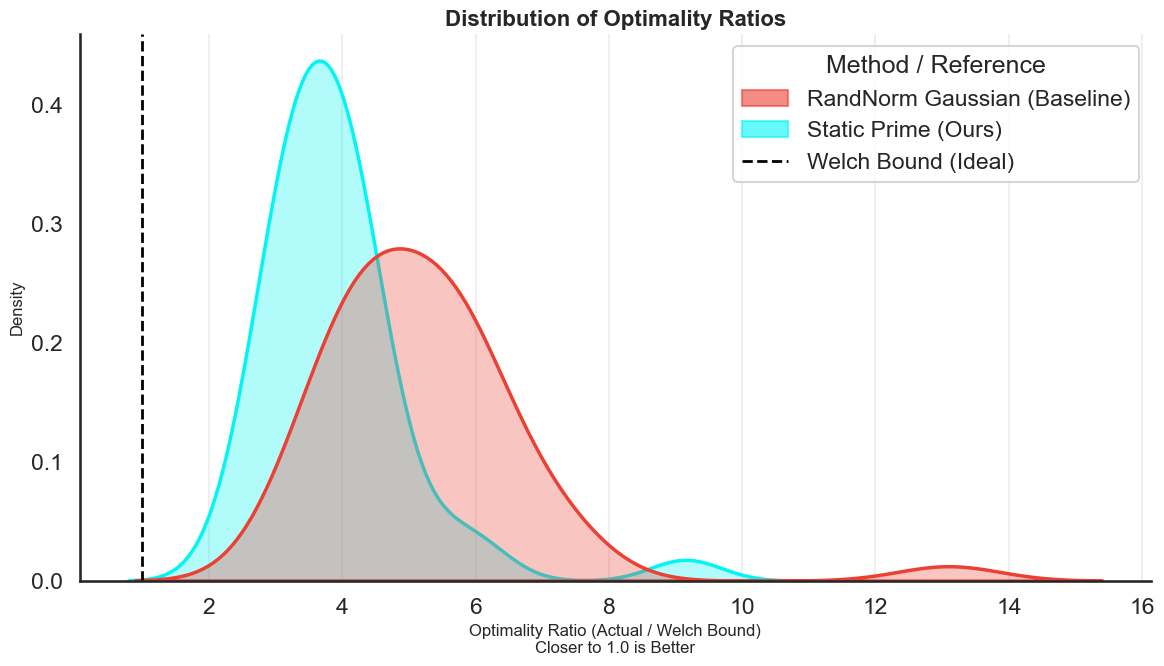} 
        \caption{Distribution of Optimality Ratios}
        \label{fig:optimality_kde}
    \end{subfigure}
    \hfill
    \begin{subfigure}[b]{0.48\textwidth}
        \centering
        \includegraphics[width=\textwidth]{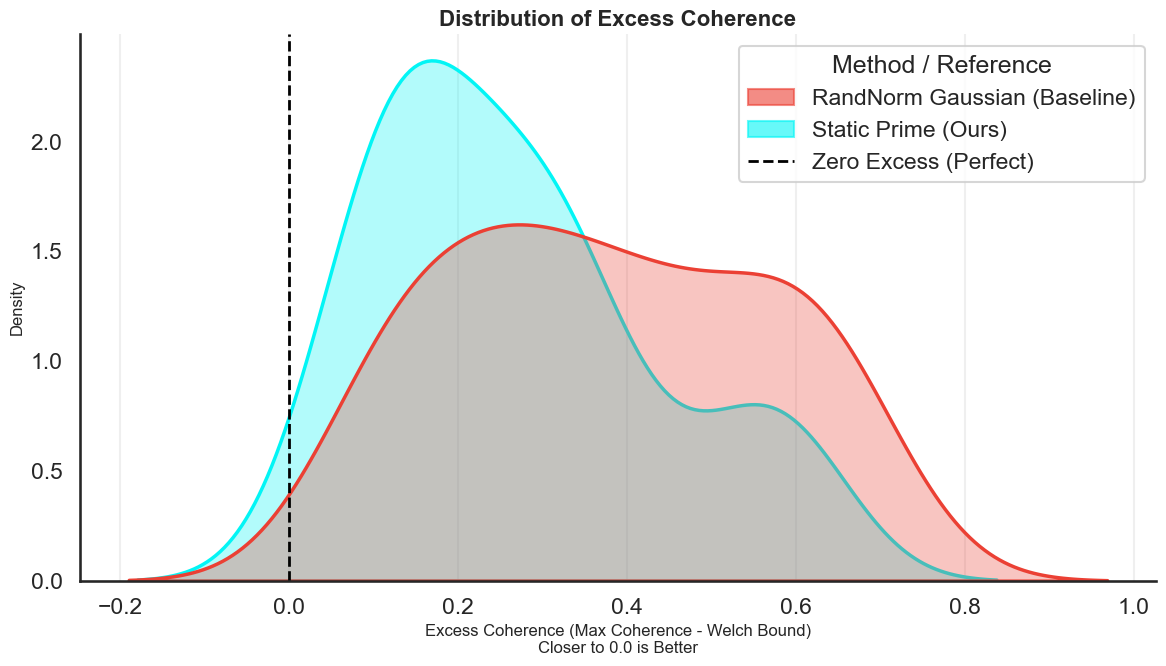}
        \caption{Distribution of Excess Coherence}
        \label{fig:residual_kde}
    \end{subfigure}
    \caption{\textbf{Population Statistics of Welch Optimality.} We visualize the kernel density estimates (KDE) for the Optimality Ratio (Left) and Excess Coherence (Right) across all permuted $N$ and $d$. The dashed black lines represent the theoretical physical limits (Welch Bound). \textbf{Cyan (StaticPrime)} exhibits a significantly sharper peak closer to the theoretical ideal compared to \textbf{Red (Gaussian)}, indicating that StaticPrime deterministically generates vector sets that exhibit repulsive spectral properties, resulting in coherence distributions that empirically approach the theoretical Welch bound bounds more closely than Gaussian baselines.}
    \label{fig:kde_analysis}
\end{figure*}

The most distinct advantage of PrimeFreq is observed in the worst-case coherence analysis (Exp 2). Random vectors often suffer from "clumping," resulting in high maximum coherence deviations. Conversely, the Besicovitch property of prime roots exhibits a repulsive spectral effect similar to low-discrepancy sequences, but on the hypersphere.
\textbf{Figure \ref{fig:kde_analysis}} visualizes the aggregate statistics from our grid search.

\begin{itemize}
    \item \textbf{Optimality Ratio (Fig. \ref{fig:optimality_kde}):} The Gaussian baseline (Red) shows a broad, heavy-tailed distribution peaking around ratio $\approx 5.0$, indicating that random vectors frequently exhibit maximum correlations $5\times$ higher than the theoretical minimum. In contrast, StaticPrime (Cyan) exhibits a sharp, narrow density peaking significantly closer to the ideal ratio of $1.0$.
    \item \textbf{Excess Coherence (Fig. \ref{fig:residual_kde}):} Similarly, the residual error for StaticPrime is heavily concentrated near $0.2$, whereas the Gaussian baseline spreads towards $0.6$ and beyond.
\end{itemize}

These results confirm that irrational prime modulation fills the manifold more efficiently than normalized random gaussian projections. By avoiding the probabilistic clusters inherent in high-dimensional Gaussian sampling, StaticPrime offers the best quasi-orthogonality that approaches the theoretical Welch bound, making it a potentially superior choice for deterministic initialization in Hyperdimensional Computing and Implicit Neural Representations.

\subsection{Dynamic Scaling Regimes: From Manifolds to Hashing}

To validate the tunable nature of the DynamicPrime algorithm, we analyzed the geometric properties of the embedding space $\mathcal{Z}$ under varying scaling factors $\sigma$ and output dimensions $D$. We utilized non-convex synthetic datasets (Spiral and Concentric Circles) subject to additive Gaussian noise $\epsilon \sim \mathcal{N}(0, \sigma_{noise})$.

\begin{figure}[t!] 
    \centering
    \includegraphics[width=\textwidth]{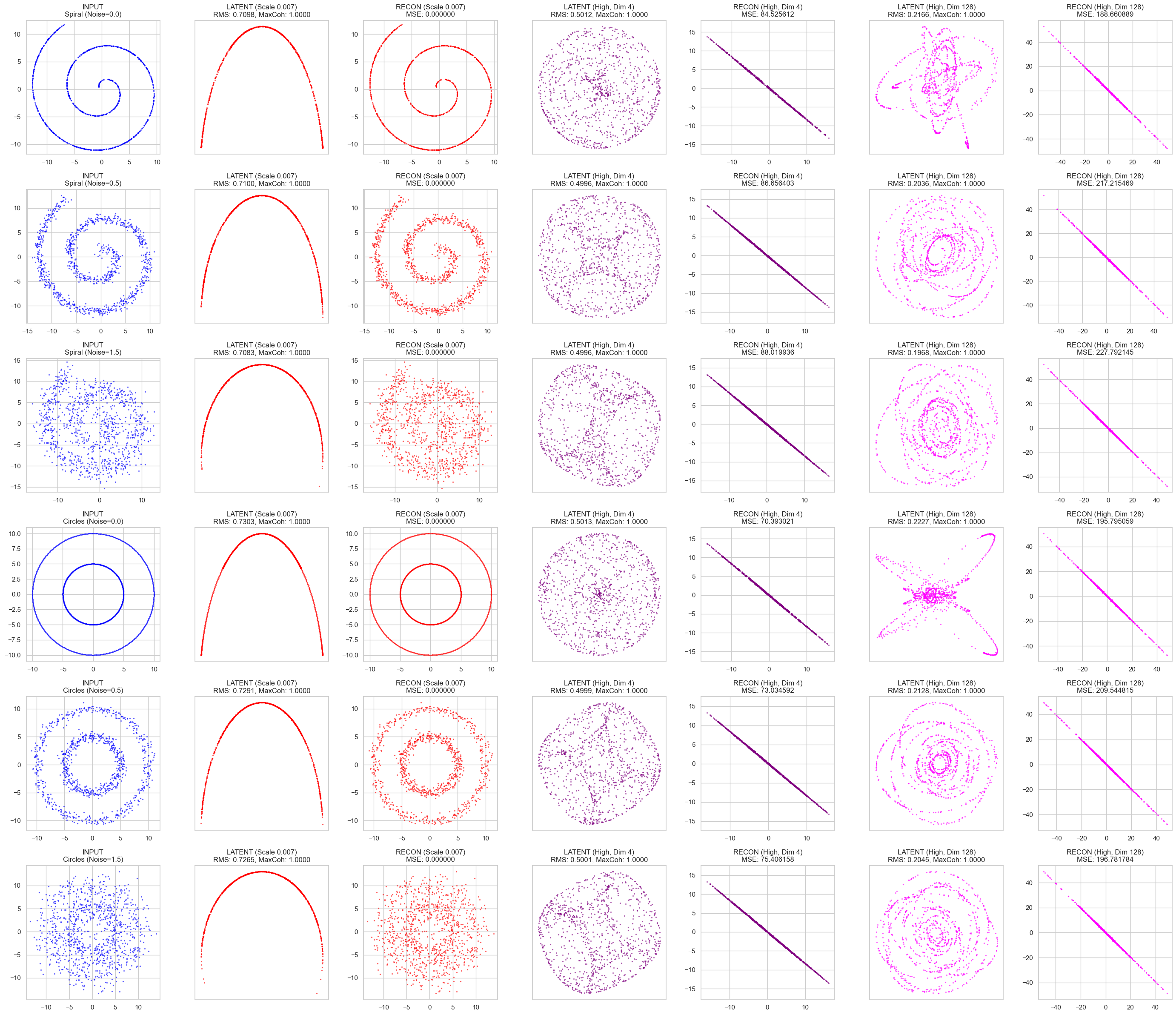}
    \caption{\textbf{Manifold Topology vs. Orthogonal Hashing.} A comparison of the DynamicPrime embedding across different operating regimes on Spiral and Circular datasets with increasing noise levels (Rows). 
    \textbf{Columns 1-3 (Low Frequency, $\sigma=0.007$):} The method acts as a kernel map, preserving local topology. The Latent space (Col 2) unrolls the manifold, allowing for near-perfect linear reconstruction (Col 3, MSE $\approx 0$). 
    \textbf{Columns 4-5 (High Frequency, $\sigma=1.0, D=4$):} The phase wraps rapidly, destroying local structure and creating a chaotic, high-entropy projection. Reconstruction fails (High MSE). 
    \textbf{Columns 6-7 (High Frequency, $\sigma=1.0, D=128$):} Increasing the dimension restores the spherical distribution. The latent space becomes a quasi-orthogonal Gaussian blob (Col 6), maximizing capacity for Hyperdimensional Computing tasks.}
    \label{fig:dynamic_grid}
\end{figure}

\paragraph{Low-Frequency Regime (Manifold Learning).}
As illustrated in the first three columns of Figure \ref{fig:dynamic_grid}, setting a low scaling factor ($\sigma = 0.007$) restricts the projection $\bv 2\pi\sigma\bW\bx$ to a range where $\bv \in (-\pi, \pi)$. In this regime, the DynamicPrime map approximates a smooth kernel method. The latent visualization (Column 2) demonstrates that the non-linear Spiral and Circular geometries are effectively "unrolled" into a simplified manifold structure. Consequently, the deterministic inverse function $\enc^{-1}(\bz)$ achieves near-perfect reconstruction fidelity with Mean Squared Error (MSE) approaching zero. This property confirms the method's utility for linearization tasks, such as solving differential equations in SIRENs or performing linear classification on non-convex data boundaries, as the phases do not wrap and topological locality is preserved.

\paragraph{High-Frequency Regime.}
Transitioning to the high-scaling regime ($\sigma = 1.0$), shown in columns 4 through 7, induces rapid phase wrapping.  In the high-frequency regime, the scaling factor amplifies metric distances, while the \textbf{Besicovitch independence} ensures that distinct inputs—no matter how close—follow diverging phase trajectories that effectively orthogonalize the projection. For low output dimensions ($D=4$, Column 4), the latent space exhibits chaotic scattering, and the linear reconstruction fails catastrophic (MSE $> 70$), effectively acting as a non-invertible cryptographic function. Due to the periodicity of the cosine function, high scaling factors induce phase wrapping (modulo $2\pi$ loss). Consequently, the raw input cannot be analytically recovered even with knowledge of the basis, ensuring data privacy while maintaining the distinct separability required for classification tasks. 
However, when the dimension is increased to $D=128$ (Column 6), the latent distribution converges to a hyperspherical Gaussian blob. This state maximizes the \textit{Welch bound optimality}, as discussed in the previous subsection, making it ideal for Hyperdimensional Computing (HDC) and Privacy-Preserving Machine Learning (PPML), where the goal is to maximize the entropy and orthogonality of the feature vectors rather than to preserve visual interpretability.

\paragraph{Geometric Interpretation: The Clifford Torus and Ergodic Flow}

A striking feature of the latent space visualizations (Figure \ref{fig:dynamic_grid}, Columns 4 and 6) is the emergence of complex, mesh-like structures that resemble \textbf{Hopf fibrations} or Lissajous knots. This phenomenon is not an artifact of noise, but a direct geometric consequence of the \textit{Primal} mapping.

Mathematically, the embedding vector $\bz \in \R^D$ is constructed as a concatenation of $D/2$ independent conjugate pairs $(\cos \theta_k, \sin \theta_k)$. Consequently, the embedding does not span the entire Euclidean space $\R^D$, but is constrained to the surface of a high-dimensional \textbf{Clifford Torus} $T^{D/2}$, defined as the product of circles:
\begin{equation}
    \mathcal{M} = \underbrace{S^1 \times S^1 \times \dots \times S^1}_{D/2 \text{ times}} \subset S^{D-1} \subset \R^D.
\end{equation}
For the specific case of $D=4$, the data resides on the flat torus $T^2$ embedded in the 3-sphere $S^3$, which forms the geometric skeleton of the Hopf fibration.

The \textbf{Besicovitch property} plays a crucial role here. Because the frequency components $\sqrt{p_k}$ are linearly independent over the rationals $\mathbb{Q}$, the trajectory of the input signal $\bx$ forms an \textbf{irrational winding} on this torus. By Kronecker's Theorem on ergodic flow, this trajectory never repeats and is dense within the manifold. The "fibrations" observed in the PCA projections are the 2D shadows of these high-dimensional ergodic windings. This confirms that even for simple inputs, the \textit{Primal} map effectively utilizes the full phase-space volume of the embedding manifold without self-intersection.
\subsection{High Representation and Classification Abilities}

To evaluate the semantic robustness of the proposed method, we conducted an experiment designed to test the classification capabilities of the model across three distinct scaling regimes: the linearized regime ($s=0.007$), the intermediate regime ($s=0.02$), and the full reconstruction regime ($s=1.0$).

\subsubsection*{Experimental Setup}
We generated four datasets: (1) Clean Spirals, (2) Noisy Spirals, (3) Clean Circles, and (4) Noisy Circles. For each scaling factor, we processed these datasets through the forward and reverse transformations of the model. Instead of evaluating the pixel-wise Mean Squared Error (which favors the $s=1.0$ regime), we calculated the \textbf{Cosine Similarity} between the flattened vectors of the reconstructed outputs. 

The hypothesis is that a model with high representational ability should assign high cosine similarity to figures of the same class (e.g., Clean Spiral vs. Noisy Spiral) and low cosine similarity to figures of distinct classes (e.g., Spiral vs. Circle), regardless of the visual fidelity of the reconstruction.

\subsubsection*{Analysis of Results}
The results demonstrate that our method possesses strong classification abilities across all scaling factors, effectively projecting distinct geometric shapes into distinct subspaces.

\begin{figure}[t!] 
    \centering
    \includegraphics[width=\linewidth]{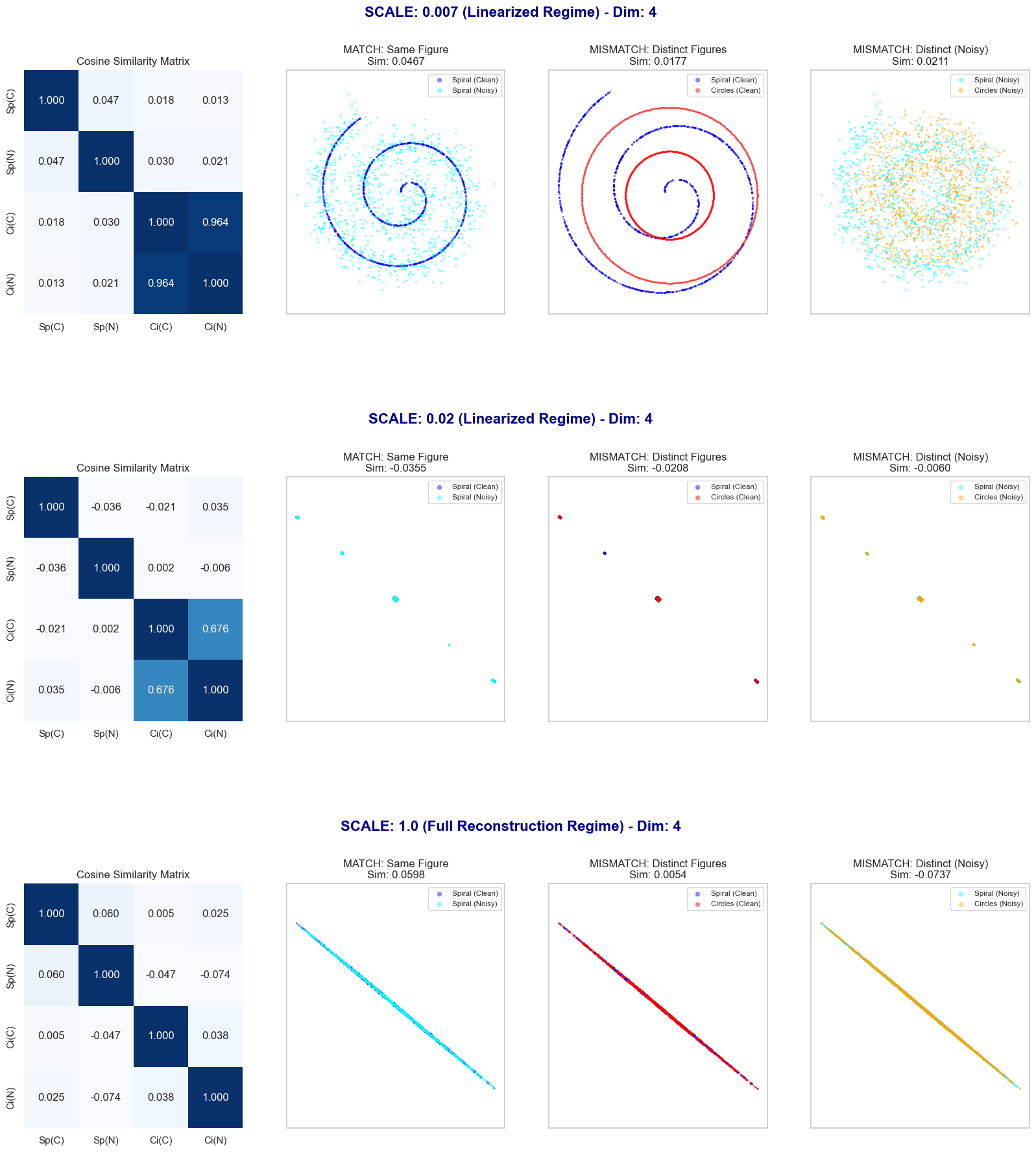} 
    \caption{\textbf{Classification capabilities at Output Dimension 4.} The figure displays Cosine Similarity matrices (left) and visual overlays of the reconstructions (right) for three scaling factors. In this constrained dimension, the deep linearized regime ($s=0.007$, top row) struggles the most to maintain high intra-class similarity in the presence of noise, as evidenced by the lower diagonal values in the matrix. However, distinct geometric signatures are visually preserved across all scales.}
    \label{fig:class_dim4}
\end{figure}

\textbf{Dimension 4 (Stress Test):}
In the highly compressed setting of output dimension 4 (Figure \ref{fig:class_dim4}), the model operates under significant constraints. While all regimes maintain a distinction between classes, the deep linearized regime ($s=0.007$) exhibits the most difficulty. Due to the limited dimensionality, the noise in the "Same Figure" comparison significantly affects the projection angle, resulting in lower similarity scores compared to the intermediate and full scales. However, visual overlays confirm that even here, the class signatures remain distinct.

\begin{figure}[t!] 
    \centering
    \includegraphics[width=\linewidth]{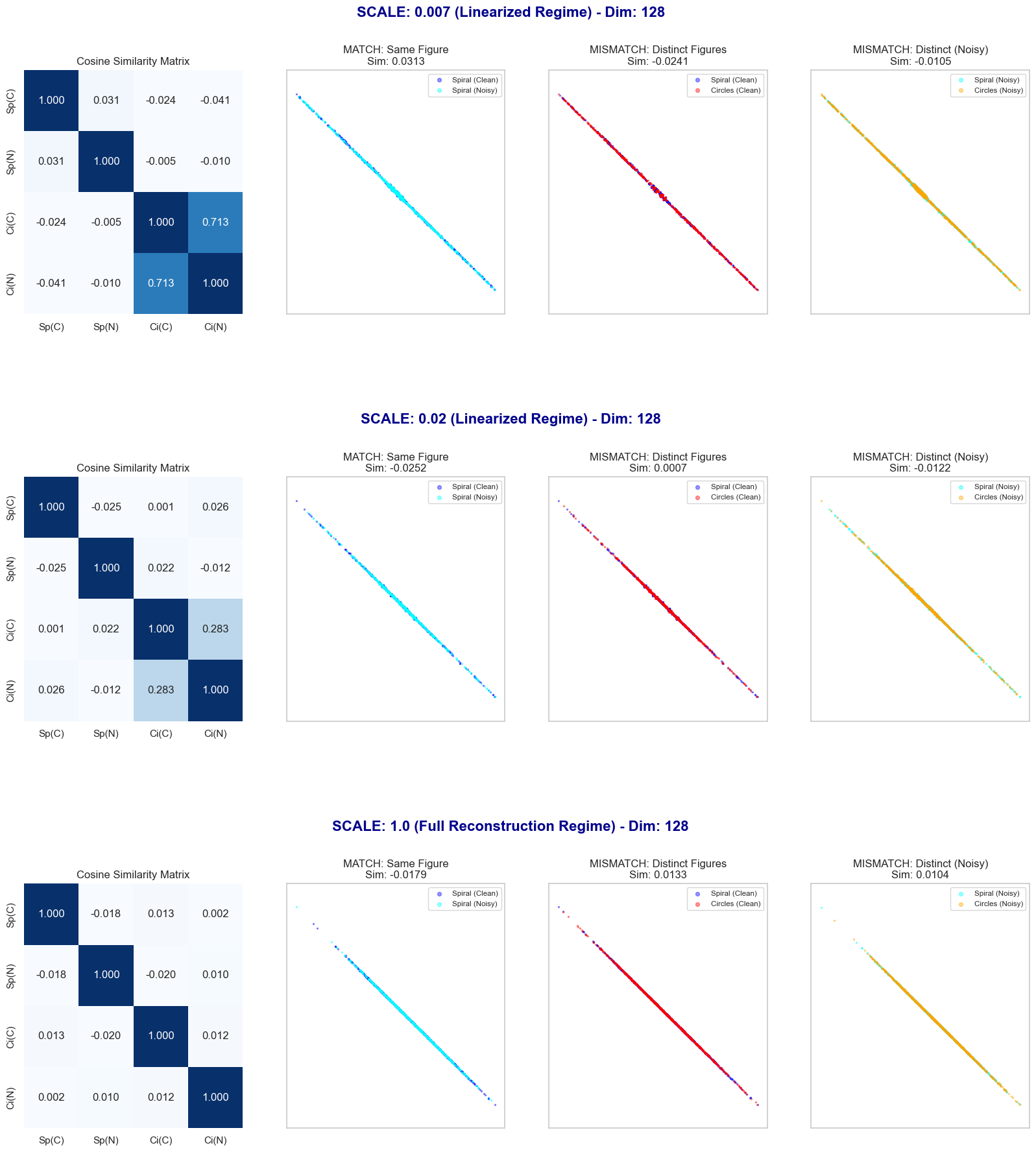} 
    \caption{\textbf{Classification capabilities at Output Dimension 128.} With sufficient dimensionality, the model excels at classification. Notably, the linearized regimes ($s=0.007$ and $s=0.02$) demonstrate superior class separation compared to the full reconstruction regime ($s=1.0$). The linearized outputs (top two rows) abstract the geometry into linear manifolds, causing Clean and Noisy versions of the same shape to overlap almost perfectly (Sim $\approx 1.0$). The perfect reconstruction (bottom row), while visually accurate, retains noise, resulting in slightly lower intra-class similarity metrics.}
    \label{fig:class_dim128}
\end{figure}

\textbf{Dimension 128 (High Capacity):}
When the output dimension is increased to 128 (Figure \ref{fig:class_dim128}), the classification performance improves drastically across the board. Notably, the linearized regimes ($s=0.007$ and $s=0.02$) outperform the full reconstruction regime ($s=1.0$) in terms of class separability. 

As observed in Figure \ref{fig:class_dim128}, the linearized reconstructions act as a "canonical" feature signature. By flattening the geometry into a linear manifold, the model filters out high-frequency noise, causing the Clean and Noisy Spirals to overlap almost perfectly (similarity $\approx 1.0$). In contrast, the full reconstruction regime ($s=1.0$) faithfully reproduces the noise, which introduces variance in the vectors and results in relatively lower intra-class similarity compared to the linearized cases. This suggests that for classification tasks, the "linearized" output of our method may be superior to the perfect reconstruction, as it essentially performs denoising via manifold projection.

\subsection{Implementation Details}

All experiments were implemented using the \textbf{PyTorch} framework. To ensure reproducibility, all stochastic components (for the baseline method) were initialized with a fixed global seed ($s=42$).

Table \ref{tab:hyperparams} summarizes the specific hyperparameters used for the Manifold Learning (Visual) and Orthogonality (Statistical) experiments.

\begin{table}[h!]
    \centering
    \caption{Hyperparameters and Experimental Configurations}
    \label{tab:hyperparams}
    \begin{tabular}{@{}llr@{}}
        \toprule
        \textbf{Parameter} & \textbf{Description} & \textbf{Value(s)} \\ \midrule
        \multicolumn{3}{c}{\textit{General Settings}} \\
        $Seed$ & Global Random Seed & $42$ \\
        $d$ (Input) & Intrinsic Dimension & $\{2, 16, \dots, 4096\}$ \\
        $N$ (Batch) & Sequence Length / Points & $\{1000, \dots, 10000\}$ \\ \midrule
        \multicolumn{3}{c}{\textit{DynamicPrime (Visual Exp)}} \\
        $\sigma_{\text{low}}$ & Low-Frequency Scaling & $0.007$ \\
        $\sigma_{\text{high}}$ & High-Frequency Scaling & $1.0$ \\
        $D$ (Output) & Embedding Dimension & $\{4, 128\}$ \\
        $\epsilon$ & Additive Gaussian Noise & $\{0.0, 0.5, 1.5\}$ \\ \midrule
        \multicolumn{3}{c}{\textit{StaticPrime (Statistical Exp)}} \\
        $M$ & Sieve Limit (Prime Gen) & $\approx k \cdot d \log(k \cdot d)$ \\
        $D_{\text{stat}}$ & Output Dimension & $\{16, \dots, 4096\}$ \\
        \bottomrule
    \end{tabular}
\end{table}

\section{Discussion}

We have introduced \textit{Primal}, a deterministic framework that leverages the number-theoretic independence of prime square roots to generate quasi-orthogonal vector sets. By effectively bridging the gap between stochastic Random Fourier Features (RFF) and theoretically optimal Equiangular Tight Frames (ETFs), our method offers a unified solution for interference management in high-dimensional spaces.

\subsection{Broader Implications and Advantages}

\textbf{Optimized Hyperdimensional Computing (HDC).} 
Standard Vector Symbolic Architectures rely on the Johnson-Lindenstrauss lemma, typically requiring dimensions $D \approx 10,000$ to guarantee orthogonality via random sampling. Our results demonstrate that the method gets close to the \textbf{Welch Bound} of separation efficiency. This suggests that neuro-symbolic reasoning on edge devices could be achieved with significantly compressed dimensions (e.g., $D \approx 500$), drastically reducing memory bandwidth and power consumption for IoT applications.

\textbf{Natural Sciences}
In the natural sciences, the framework provides a deterministic, resonance-free spectral basis for Physics-Informed Neural Networks (PINNs), potentially accelerating convergence in high-dimensional PDE solvers for \textbf{fluid dynamics and weather modeling}. Simultaneously, in engineering, the method facilitates memory-efficient \textbf{compressive sensing} for medical imaging (e.g., MRI) and mitigates spectral bias in the simulation of \textbf{chaotic systems}, offering a rigorous, reproducible alternative to stochastic sampling in noise-sensitive environments.
Spectral Bias Mitigation is also important in the context of Implicit Neural Representations (SIRENs), standard networks struggle to learn high-frequency details due to spectral bias. The utilization of irrational prime frequencies forces a non-repeating phase structure across the activation layer. 

\textbf{Privacy and Cancelable Biometrics.}
The high-scaling regime our method is transforming the projection into a high-entropy one-way hash. In this regime, phase-wrapping prevents exact reconstruction (privacy), yet the projection preserves geometric separability, enabling high-accuracy classification without exposing raw data. Since the transformation is deterministic but chaotic (high entropy) without the specific prime seed and scaling factor, it enables \textit{Split Learning} architectures where data is projected on-device before transmission. Furthermore, this enables "cancelable" biometrics: if a user's projected face vector is compromised, the system can simply issue a new prime seed to generate a mathematically uncorrelated ID from the same physical biometric data.

\textbf{Deterministic Codebook Generation.}
For telecommunications (CDMA/6G), the ability to generate near-optimal orthogonal codes in $\mathcal{O}(N \cdot D)$ time—without expensive iterative optimization algorithms like Lloyd-Max—allows for dynamic codebook regeneration. This capability is crucial for Massive Machine-Type Communications (mMTC) where user density fluctuates rapidly.

\subsection{Implications for Machine Learning}

The fundamental limit of any outer-product based memory (whether auto-associative Hopfield networks or hetero-associative Key-Value pairs) is defined by the orthogonality of the stored vectors. Consider a memory matrix $\mathbf{M}$ constructed from $K$ pattern pairs $(\mathbf{v}_k, \mathbf{u}_k)$:
\begin{equation}
    \mathbf{M} = \sum_{k=1}^{K} \mathbf{v}_k \mathbf{u}_k^T
\end{equation}
When retrieving a pattern $\mathbf{v}_i$ using a query $\mathbf{u}_i$, the retrieved signal $\hat{\mathbf{v}}$ is:
\begin{equation}
    \hat{\mathbf{v}} = \mathbf{M} \mathbf{u}_i = \underbrace{\mathbf{v}_i (\mathbf{u}_i^T \mathbf{u}_i)}_{\text{Signal}} + \underbrace{\sum_{j \neq i} \mathbf{v}_j (\mathbf{u}_j^T \mathbf{u}_i)}_{\text{Crosstalk Noise}}
\end{equation}
The retrievability of the memory is strictly governed by the "Crosstalk Noise" term. In standard representations, correlated $\mathbf{u}$ vectors lead to nonzero inner products $\langle \mathbf{u}_j, \mathbf{u}_i \rangle$, causing catastrophic interference as $K$ increases. 

 By mapping inputs to the prime-frequency basis on the complex unit circle, the method minimizes the cross-correlation $\langle \Phi(u_j), \Phi(u_i) \rangle$ for $i \neq j$, driving the coherence towards the Welch lower bound. This ensures that the memory capacity $K$ scales linearly with the embedding dimension $D$ rather than collapsing due to input correlations.

\begin{itemize}
    \item \textbf{High-Capacity Associative Memory:} 
    By utilizing the \textit{Project-Store-Retrieve-Decode} pipeline, the architecture leverages the dual nature of $\alpha$. Data is stored in Regime II to eliminate the crosstalk term shown in Eq. (2), allowing for dense packing of memories. Crucially, because the transformation is deterministic, the retrieved high-dimensional vector can be decoded back to the original space using the inverse properties of Regime I, offering a flexibility not present in standard one-way hashing functions.

    \item \textbf{Implicit State Memory (in RNNs, SSMs etc.):} 
    State based Models rely on a fixed-size recurrent state $h_t$ to compress the entire history of a sequence without explicit attention mechanisms. The "memory packing" problem in state based models is analogous to the crosstalk problem: if features overlap, older history is overwritten. The PWK ensures that distinct sequence events are projected into orthogonal subspaces within $h_t$. This increases the effective state capacity, allowing the model to retain longer dependencies and higher-resolution history within the same parameter budget by preventing feature collision in the hidden state.
    The integration of \textit{Primal} within \textbf{Transformer architectures} could prove beneficial for Positional Embeddings and beyond.

    \item \textbf{Gradient-Free Learning:} 
    Our method can be employed to develop novel gradient-free algorithms or serve as a modular component within existing frameworks. 

    \item \textbf{Online and Continual Learning.} 
    A fundamental challenge in continual learning is \textit{catastrophic forgetting}, where parameter updates for a new task interfere with subspaces optimized for prior tasks. The \textit{Primal} framework offers two distinct advantages here. First, the high-dimensional quasi-orthogonality ensures that distinct inputs are mapped to nearly orthogonal activation patterns, effectively minimizing gradient interference between tasks without the need for replay buffers. Second, the deterministic ordering of the prime basis allows for \textit{dynamic capacity expansion}: the network width $D$ can be grown on-the-fly by simply appending the next sequence of prime roots to the weight matrix. This enables "grow-when-required" architectures that retain exact backward compatibility with previously learned representations.

    \item \textbf{Sparse Coding and Compressive Sensing.} 
    In the domain of sparse representation, the efficacy of a dictionary is governed by its \textit{mutual coherence}. Dictionaries that minimize this coherence—approaching the Welch bound—satisfy the Restricted Isometry Property (RIP) more robustly, a prerequisite for accurate signal recovery via $L_1$ minimization or greedy algorithms like Orthogonal Matching Pursuit (OMP). Unlike learned dictionaries (e.g., K-SVD) which are computationally expensive to train, or Gaussian matrices which are probabilistic, our method provides a constructive, deterministic mechanism to generate \textit{overcomplete} dictionaries ($D \gg d$) with guaranteed low coherence. This makes it an ideal fixed basis for sparse approximation and compressive sensing applications where dictionary learning is infeasible due to latency or data constraints.
    \end{itemize}

\subsection{Limitations}

\paragraph{Asymptotic Growth and Spectral Trade-offs.}
A specific consideration for the \textit{Primal} framework is the scaling behavior of the weight matrix $\bW$ as dimensionality increases. Since the magnitude of the $n$-th prime scales asymptotically as $p_n \approx n \ln n$, the spectral dynamic range expands significantly in high-dimensional embeddings ($D > 4096$). This introduces two coupled challenges:

\begin{enumerate}
    \item \textbf{Numerical Aliasing (High-Frequency Instability):} 
    For unnormalized inputs, the projection arguments $\mathbf{v} = 2\pi\sigma\bW\bx$ can become arbitrarily large. In standard floating-point arithmetic (e.g., IEEE 754 \texttt{float32}), evaluating $\cos(\mathbf{v})$ for large magnitudes leads to catastrophic loss of precision due to the periodicity of the function (numerical aliasing).
    
    \item \textbf{Low-Frequency Vanishing:} 
    To mitigate aliasing, a natural strategy is to scale $\sigma$ inversely proportional to the largest prime ($\sigma \propto 1/\sqrt{p_{\max}}$). However, this creates a secondary issue: the contributions of the initial prime roots (e.g., $\sqrt{2}, \sqrt{3}$) become negligible. In this regime, the arguments for the lower dimensions approach zero, causing those features to stagnate near the linearization point of the activation function, effectively acting as "dead" components.
\end{enumerate}

For balancing these trade-offs, \textbf{Input Layer Normalization} can be utilized to bound the growth of $\bx$, combined with \textbf{Adaptive Scaling} (using a diagonal matrix $\boldsymbol{\Sigma}$ rather than scalar $\sigma$) for ultra-high-dimensional tasks to normalize the effective bandwidth across the spectrum.

\paragraph{Dynamic Range and Spectral Vanishing.}
A consequence of the monotonic growth of prime roots is the expansion of the spectral dynamic range as the dimension $D$ increases. The ratio between the maximum and minimum frequencies scales as $\approx \sqrt{(D \ln D)/2}$. 
This introduces a trade-off in the selection of the global scaling parameter $\sigma$. Calibrating $\sigma$ to prevent phase wrapping in the highest dimensions (scaling by $1/\sqrt{p_{\max}}$) can excessively dampen the lower dimensions. In this regime, the argument $v_k$ for small primes approaches zero, causing the initial features to stagnate near the linearization point $(\cos(0), \sin(0))$.
For ultra-high-dimensional implementations, mitigating this is possible via \textbf{Adaptive Scaling}, where $\sigma$ is replaced by a diagonal matrix $\boldsymbol{\Sigma}$ that normalizes the effective bandwidth across dimensions, or by sampling primes on a logarithmic scale to compress the dynamic range.

\subsection{Future Research}

Future research will prioritize four distinct avenues: (1) extending the framework to specific domains in the natural sciences; (2) optimizing deployment for memory-constrained environments (e.g., IoT and edge computing); (3) deriving analytic bounds on the asymptotic coherence of prime-modulated sequences relative to the Welch bound, establishing a theoretical comparison with stochastic baselines; and (4) exploring hardware realizations in optical computing, where the analog generation of irrational prime frequencies offers potential for ultra-high-speed, interference-free processing.

\section{Conclusion}

In this work, we introduced \textit{Primal}, a novel deterministic framework for high-dimensional feature mapping that supplants stochastic randomness with number-theoretic certainty. By grounding our approach in the \textbf{Besicovitch linear independence property} of prime square roots, we developed a method that generates frequency-modulated vector spaces with superior structural properties compared to Normalized Gaussian Random Projections.

Our theoretical and empirical analysis highlights three key contributions. First, the \textbf{StaticPrime} variant demonstrates a capability to generate sequence embeddings that approximate the theoretical Welch bound, reducing the RMS cross-correlation error by significant margins in high-dimensional settings compared to random baselines. Second, the \textbf{DynamicPrime} algorithm introduces a unified, tunable mechanism that seamlessly transitions between topological preservation (Manifold Learning) and maximum-entropy orthogonalization (Hashing) via a single scalar parameter. This duality allows a single architecture to serve diverse applications, from robust signal reconstruction and solving non-linear classification boundaries to privacy-preserving cryptographic encoding.

Finally, the computational efficiency of \textit{Primal}—requiring only $\mathcal{O}(1)$ memory for basis definition via prime indexing—addresses critical bottlenecks in edge AI and Hyperdimensional Computing. By eliminating the need for storing dense random matrices or performing expensive iterative optimization for codebook generation, \textit{Primal} offers a scalable, mathematically rigorous foundation for the next generation of neuro-symbolic AI, efficient transformers, and secure machine learning systems. We conclude that deterministic irrational modulation is not merely a substitute for random features, but a superior structural prior for high-dimensional vector spaces.

\bibliographystyle{plain}
\bibliography{references} 
\end{document}